\documentclass[runningheads]{llncs}

\usepackage{eccv}

\usepackage{multirow}
\usepackage{eccvabbrv}

\usepackage{graphicx}
\usepackage{booktabs}

\usepackage[accsupp]{axessibility}  

\usepackage[pagebackref,breaklinks,colorlinks,citecolor=eccvblue]{hyperref}

\usepackage{orcidlink}

\begin{document}

\title{DAMSDet: Dynamic Adaptive Multispectral Detection Transformer with Competitive Query Selection and Adaptive Feature Fusion} 

\titlerunning{DAMSDet}

\author{Junjie Guo\inst{1} \and
Chenqiang Gao\inst{2*} \and
Fangcen Liu\inst{1} \and
Deyu Meng\inst{3} \and
Xinbo Gao\inst{1}}

\authorrunning{Guo et al.}

\institute{School of Communications and Information Engineering, Chongqing University of Posts and Telecommunications, Chongqing 400065, China
 \and
 School of Intelligent Systems Engineering, Sun Yat-sen University, Shenzhen, Guangdong 518107, China.
 \and
 Deyu Meng is with the School of Mathematics and Statistics, Xi'an Jiaotong University, Xi'an, Shanxi, 710049, China.}

\maketitle

\begin{abstract}
  Infrared-visible object detection aims to achieve robust even full-day object detection by fusing the complementary information of infrared and visible images. However, highly dynamically variable complementary characteristics and commonly existing modality misalignment make the fusion of complementary information difficult. 
  In this paper, we propose a Dynamic Adaptive Multispectral Detection Transformer (DAMSDet) to simultaneously address these two challenges. 
  Specifically, we propose a Modality Competitive Query Selection strategy to provide useful prior information. This strategy can dynamically select basic salient modality feature representation for each object. To effectively mine the complementary information and adapt to misalignment situations, we propose a Multispectral Deformable Cross-attention module to adaptively sample and aggregate multi-semantic level features of infrared and visible images for each object. In addition, we further adopt the cascade structure of DETR to better mine complementary information.
  Experiments on four public datasets of different scenes demonstrate significant improvements compared to other state-of-the-art methods. The code will be released at \url{https://github.com/gjj45/DAMSDet}. 
\end{abstract}

\section{Introduction}
\label{sec:intro}

\quad 
Object detection is a fundamental task in computer vision and most of research works are based on visible images with detailed object information, e.g., texture and color information.
Thanks to the development of deep learning, the object detection technique has made great progress.
However, it is still challenged by poor imaging conditions, such as low illumination, smoke, fog, and so on, which could make objects pretty obscure and further obviously degrade the performance of object detection.  
\begin{figure}  
    \centering
    \includegraphics[trim=44mm 23mm 48mm 20mm,clip,width=0.5\textwidth]{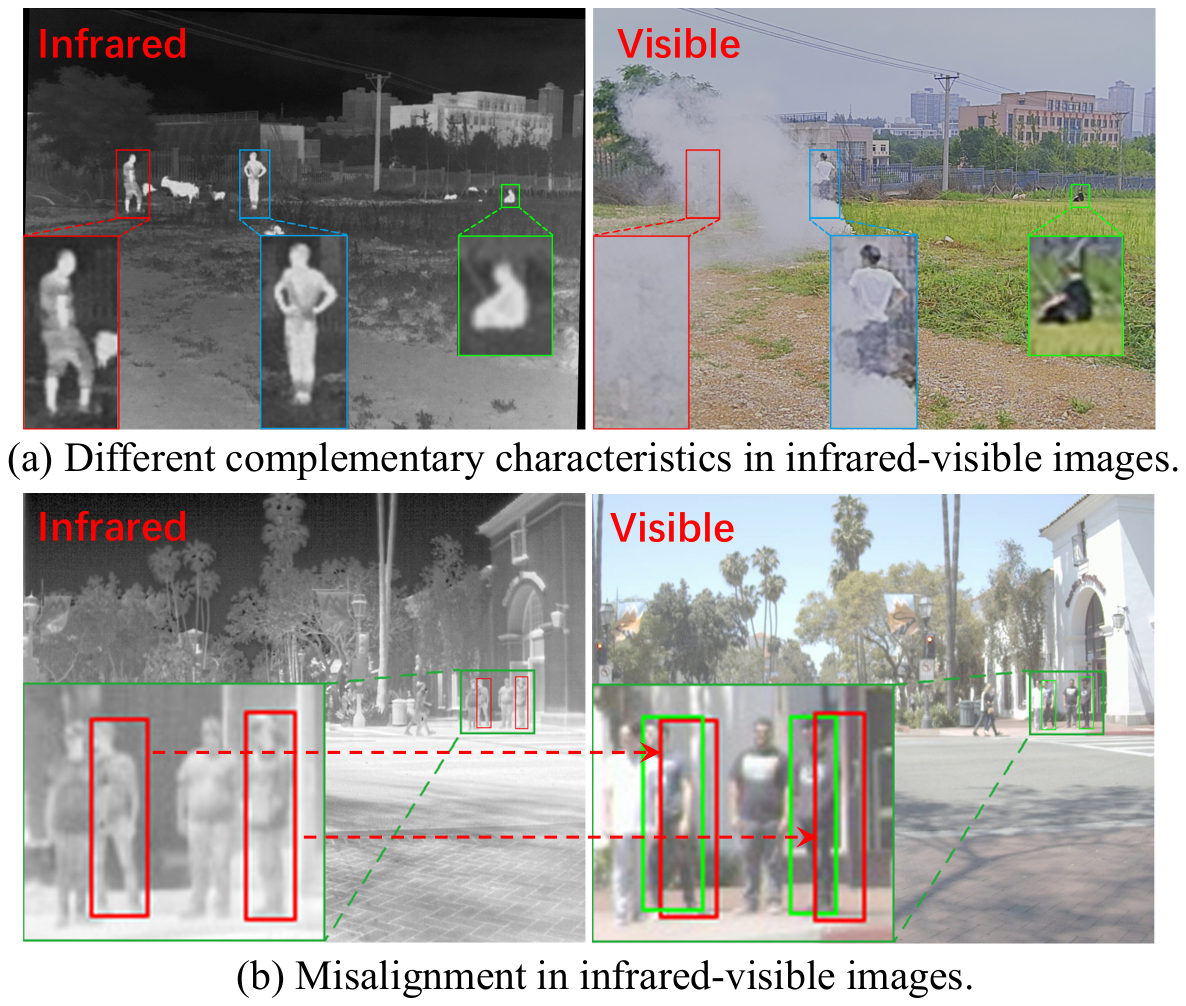}
    \caption{Illustrations of two typical challenges in infrared-visible object detection. (a) Three pedestrians represent different complex complementary characteristics. In this example, the objects in the visible image provide unuseful interference information (red), partial complementary information (blue), and full complementary information (green).  (b) One example of the misalignment problem, in which the ground truths of infrared and visible objects appear obvious dislocation. This misalignment commonly happens in infrared-visible images.  We propose a Multispectral Transformer Decoder with Multispectral Deformable Cross-attention module to simultaneously address these two typical challenges.}
    \label{fig:fig1_challenge}
    \vspace{-10pt}
\end{figure}
Thus, infrared images are introduced into the object detection task. 
Different from visible imaging, infrared imaging captures the thermal radiation of objects, making it unaffected by illumination, smoke and fog occlusion conditions. 
Therefore, infrared imaging can still well capture objects even in low illumination, heavy smoke, or fog, whereas the detailed texture and color information will be lost.
These complementary characteristics of infrared and visible imaging not only can improve the performance of object detection, but also is considered to be promising to implement full-day object detection. 
Thus, infrared-visible object detection has attracted extensive attention in recent years \cite{zhou2020improving,zhang2021weakly,fu2023lraf,shen2024icafusion,li2023stabilizing,yang2022baanet}.  

However, existing methods tend to neglect the modality interference encountered in complex scenes during the fusion process.
For the case that the object signal in one modality is poor or absent, directly fusing the information of two modalities will bring in unuseful interference information, which could lead to feature confusion and thus degrade the object detection performance.
For example, as shown in Fig. \ref{fig:fig1_challenge}\hyperref[fig:fig1_challenge]{(a)}, the pedestrian within the smoke fully disappears, and intuitively, the best way is to suppress or discard the visible information for that pedestrian. 
Some works learn a global fusion weight to adapt to specific scenes, the representative ones of which are to adopt the illumination-aware network to obtain illumination score as the global fusion weight \cite{li2019illumination, zhou2020improving, yang2022baanet}.
Other works learn local region fusion weights through bounding-box level semantic segmentation \cite{li2019illumination, zhou2020improving, yang2022baanet}, or regions of interest (ROI) prediction \cite{zhang2019weakly,zhang2021weakly,kim2021uncertainty}.

Actually, due to the fully different imaging principles, the complementary characteristics in infrared-visible images appear highly variable with the specific scenes and objects, as shown as Fig. \ref{fig:fig1_challenge}.
Especially, from Fig. \ref{fig:fig1_challenge}\hyperref[fig:fig1_challenge]{(a)}, we can observe that three pedestrians have obviously different complementary characteristics. 
The one with the green bounding box has good complementary information in both modalities, while the one with the red bounding box has only infrared information available, as mentioned previously.
In contrast, the one with the blue bounding box has partial information available in both modalities, which commonly exists in practical applications. 
This situation would make current methods fail to effectively fuse features, even for above mentioned region-based weight fusion methods in which the segmented or predicted regions are usually bigger than objects.
Therefore, more fine-grained two-modality information fusion still remains a challenge.

Another important challenge in infrared-visible object detection is the modality misalignment problem. 
Most feature fusion methods assume that the two modalities are well-aligned.
However, precise registration is difficult because infrared-visible images often exhibit significant visual differences and are not always captured at the exact same timestamp \cite{zhu2023multi}.
As a result, even through manual registration, the imaging objects in two modalities for the same one are usually misaligned, as shown in Fig. \ref{fig:fig1_challenge}\hyperref[fig:fig1_challenge]{(b)}.
This could lead to disrupting the consistency of fused feature representation of current methods, affecting the final detection performance. 
AR-CNN \cite{zhang2019weakly,zhang2021weakly} explicitly learned the offsets of objects in both modalities to achieve alignment on object features. 
However, this method requires additional paired bounding-box annotations of two modalities during training, which is time-consuming and labor-intensive.

In this paper, we propose a novel adaptive infrared-visible object detection method, which
contains a Multispectral Transformer Decoder with Multispectral Deformable Cross-attention module to simultaneously address the above two challenges inspired by the deformable cross attention \cite{zhu2020deformable}. Specifically, we adopt an effective strategy of adaptive sparse feature sampling and weight aggregation on two-modality feature maps of different semantic levels. This strategy can effectively fuse fine-grained complementary information even when two modalities are misaligned. Since the two challenges of fine-grained information fusion and modality alignment are simultaneously handled in a single module, our method is more efficient than existing methods which usually handle them separately.
Furthermore, unlike the one-step fusion strategy adopted by existing methods, the information fusion of each specific object in our method happens on different semantic levels, which makes the complementary information fully be mined and utilized. Actually,  we observed that the complementary information of two modalities also dynamically varies with the semantic levels, as discussed in Sec. \ref{decoder}. This is similar to our observation on scenes and objects discussed previously. Thus, our adaptive multi-level fusion is more reasonable.

In order to provide reliable input at the early stage, we design a Competitive Query selection strategy to select dominant modality features for each object as initial position and content queries to build a basic salient feature representation for the Multispectral Transformer Decoder, which can provide useful prior information for following processing. To further exploit more reliable and comprehensive complementary information step by step, the cascaded layer structure of DETR \cite{zhang2022dino} is employed in this paper. Totally, our method is similar to the human observation pattern which dynamically focuses on objects in each modality and gradually aggregates key information of two modalities. 

Our contributions can be summarized as follows:

\begin{itemize}
    \item[$\bullet$]  We propose a novel infrared-visible object detection method, named DAMSDet, which can dynamically focus on dominant modality objects and adaptively fuse complementary information.
    \item[$\bullet$]  We propose a Competitive Selection strategy for multimodal initialization queries to dynamically focus on the dominant modality of each object and provide useful prior information for following fusion process.
    \item[$\bullet$]  We propose a Multispectral Deformable Cross-attention module, which can simultaneously adaptively mine fine-grained partial complementary information at different semantic levels and adapt to modality misalignment situations.
    \item[$\bullet$]  Experiments on four public datasets with different scenarios demonstrate that the proposed method achieves significant improvement compared with other state-of-the-art methods.
\end{itemize}

\vspace{-10pt}
\section{Related Work}
\vspace{-5pt}

\textbf{Infrared-Visible object detection.}
Previous research in infrared-visible object detection are primarily built upon the single modality object detection frameworks, which are generally divided into one-stage object detectors, such as Faster RCNN \cite{ren2015faster}, and two-stage object detectors, such as YOLO \cite{redmon2016you,redmon2017yolo9000,redmon2018yolov3,wang2023yolov7}.

In order to fuse the complementary information of infrared and visible images,
Konig \textit{et al.} \cite{konig2017fully} introduced a fully convolutional fusion RPN network, which fused infrared and visible image features by concatenation, and concluded that halfway fusion can obtain better result \cite{liu2016multispectral}. 
On this foundation, some studies designed CNN-based attention modules to better exploit the potential complementation of infrared and visible images \cite{qingyun2022cross,roszyk2022adopting,cao2023multimodal}. 
Additionally, other works introduced transformer-based fusion modules to capture a more global complementary relationship between infrared and visible images \cite{qingyun2021cross,fu2023lraf,zhu2023multi,shen2024icafusion}.

In addition to the above methods of directly fusing image features. Some works adopted illumination information as global weights to fuse infrared and visible image features or post-fuse the multibrance detection results to reduce the impact of interfering information \cite{li2019illumination,zhou2020improving,yang2022baanet}.
Considering that the complementary characteristics of different regions could be different, some studies introduced bounding box level semantic segmentation \cite{li2018multispectral,li2019illumination,cao2019box,zhang2020multispectral,zhang2021guided} or regions of interest (ROI) prediction \cite{zhang2019weakly,zhang2021weakly,kim2021uncertainty} to guide the fusion of different regions. Other works also utilized the confidence or uncertainty scores of regions to post-fuse the predictions of multibranchs \cite{li2022confidence,li2023stabilizing}.

To address the challenge of modality misalignment, Zhang \textit{et al.} \cite{zhang2019weakly,zhang2021weakly} developed the AR-CNN network and explicitly aligned the features of two modalities by incorporating additional paired bounding box annotations to learn object misalignment. Kim \textit{et al.} \cite{kim2021mlpd} also employed a multi-label learning approach to adapt object detection in scenes with misalignment.

The above methods significantly enhance the performance of infrared-visible object detection.
However, these methods perform overall image feature fusion or one-step reweighting region feature fusion, making it difficult to mine complete complementary information in complex scenes.
In contrast, our proposed method links the misalignment problem with complementary feature fusion, which gradually and adaptively mines object-specific fine-grained complementary information at multiple semantic levels.

\textbf{End to end Object Detectors.}
In recent years, Carion \textit{et al.} \cite{carion2020end} first introduced the end-to-end object detector based on transformer called DEtection TRansformer(DETR).  
It views object detection as a set prediction problem and uses binary matching to directly predict one-to-one object sets during training.
This greatly simplifies the object detection pipeline, eliminating the need for manual anchor box design or post-processing with NMS. 
Although DETR has these advantages, it has the problem of slow training convergence and many DETR variants have been proposed to address this issue. 
Deformable DETR \cite{zhu2020deformable} accelerated training convergence by predicting 2D anchor points and designing a Deformable cross-attention module to sparsely sample features around reference points.
Conditional DETR \cite{meng2021conditional} decoupled the content and position operations and proposes conditional cross-attention to accelerate training convergence.
Efficient DETR \cite{yao2021efficient} proposed a more efficient pipeline by combining dense detection and sparse set detection.
DAB-DETR \cite{liu2022dab} introduced 4D reference points to optimize the prediction box layer by layer.
DN-DETR \cite{li2022dn} accelerated the training process and label-matching effect by introducing query denoising in the training phase.
DINO \cite{zhang2022dino} integrated the above works to build a powerful DETR detection framework.
For detection efficiency, RT-DETR \cite{lv2023detrs} constructed a real-time and end-to-end object detector by designing an Efficient Hybrid Encoder and adopting an IoU-aware Query Selection strategy.

Recently, a study on multispectral pedestrian detection based on DETR was conducted, which designed three prediction branches and an instance-aware modality-balance loss to align the contributions of each modality \cite{xing2023multispectral}. 
In contrast, our method has only one prediction branch and guides the feature fusion of each specific object through a dynamic Modality Competitive Query Selection strategy.  This dynamic guidance strategy based on sample changes could be more effective for infrared-visible object detection in complex and changeable scenes.

\vspace{-10pt}
\section{Method}
\label{sec:method}
\vspace{-5pt}
\subsection{Overview}
\begin{figure}
    \centering
    \includegraphics[trim=2mm 66mm 2mm 65mm,clip,width=\textwidth]{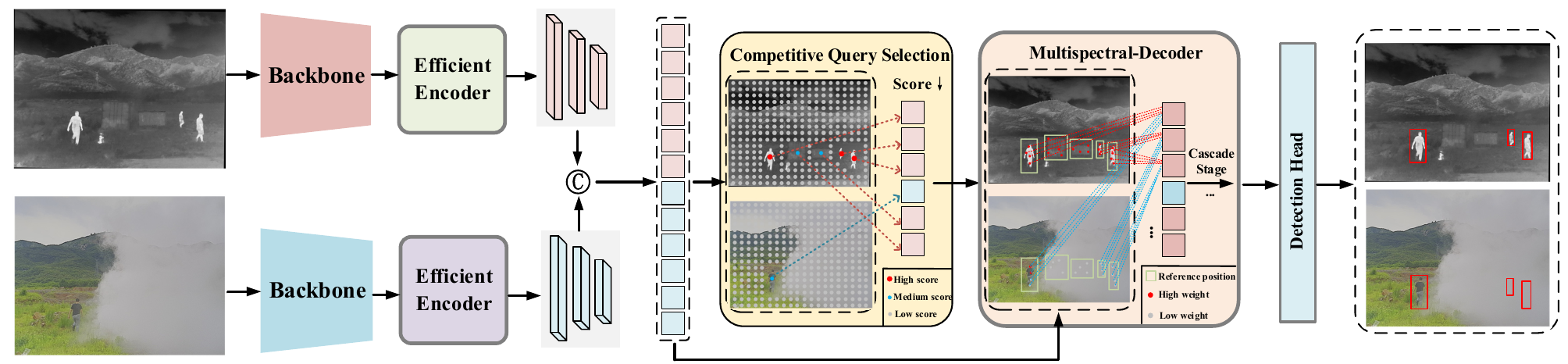}
    \caption{Overview of DAMSDet. Our DAMSDet comprises four main components: two modality-specific CNN backbones to extract features, two modality-specific Efficient Encoders \cite{lv2023detrs} to encode features, a Modality Competitive Query Selection module to select initial object queries, and a Multispectral Transformer Decoder to mine complementary information and refine queries. }
    \label{fig:ms_detr_overview}
    \vspace{-8pt}
\end{figure}

\quad  The overview of our DAMSDet is illustrated in Fig. \ref{fig:ms_detr_overview}. Our method contains four main components: two modality-specific CNN backbones, two modality-specific Efficient Encoders, Modality Competitive Query Selection, and a Multispectral Transformer Decoder.
Given a pair of infrared and visible images, we first extract and encode their features separately using two modality-specific CNN backbones and two modality-specific Efficient Encoders.  
Subsequently, the encoded features are flattened, concatenated, and input into the Modality Competitive Query Selection module. This module selects salient modality features to serve as the initial object queries. Next, these modality-specific object queries enter the Multispectral Transformer decoder, which refines them with multiple semantic levels of infrared and visible feature maps through cascaded decoder layers. Finally, these refined object queries are mapped through the detection head to obtain the bounding boxes and classification scores of all objects.

The Efficient Encoder combines Transformer and CNN to significantly reduce computational complexity, following the structure of RT-DETR \cite{lv2023detrs}.
In the following, we will elaborate on the proposed Modality Competitive Query Selection strategy and Multispectral Transformer Decoder with the Multispectral deformable cross-attention module in detail.

\vspace{-5pt}
\subsection{Modality Competitive Query Selection}
\quad The object queries in DETR are a set of learnable embeddings, which contain the content and position information of the objects.
These queries serve as object feature representations that interact with the image feature sequence in the decoder and generate bounding boxes and classification scores through the prediction head mapping.
In addition to setting object queries as learnable embeddings, there are also some methods to use the Top-$K$ score features as the initial object query \cite{zhu2020deformable,zhang2022dino,yao2021efficient}. 
Learnable object queries are difficult to optimize because they have no explicit physical meaning \cite{lv2023detrs}.
In infrared and visible images, a gap exists between two modality features, further complicating the optimization of learnable object queries. 
Therefore, selecting object queries from the encoded feature maps is more suitable for the dynamic changes of complementary characteristics in the infrared-visible object detection task. 

Concretely, we concatenate encoded feature sequences from the infrared and visible modalities and feed them into a linear projection layer to obtain feature point scores.
From this combined feature representation, we select the Top-$K$ scoring features as the initial object queries. 
\begin{figure}
    \centering
    \includegraphics[trim=10mm 50mm 10mm 40mm,clip,width=0.7\textwidth]{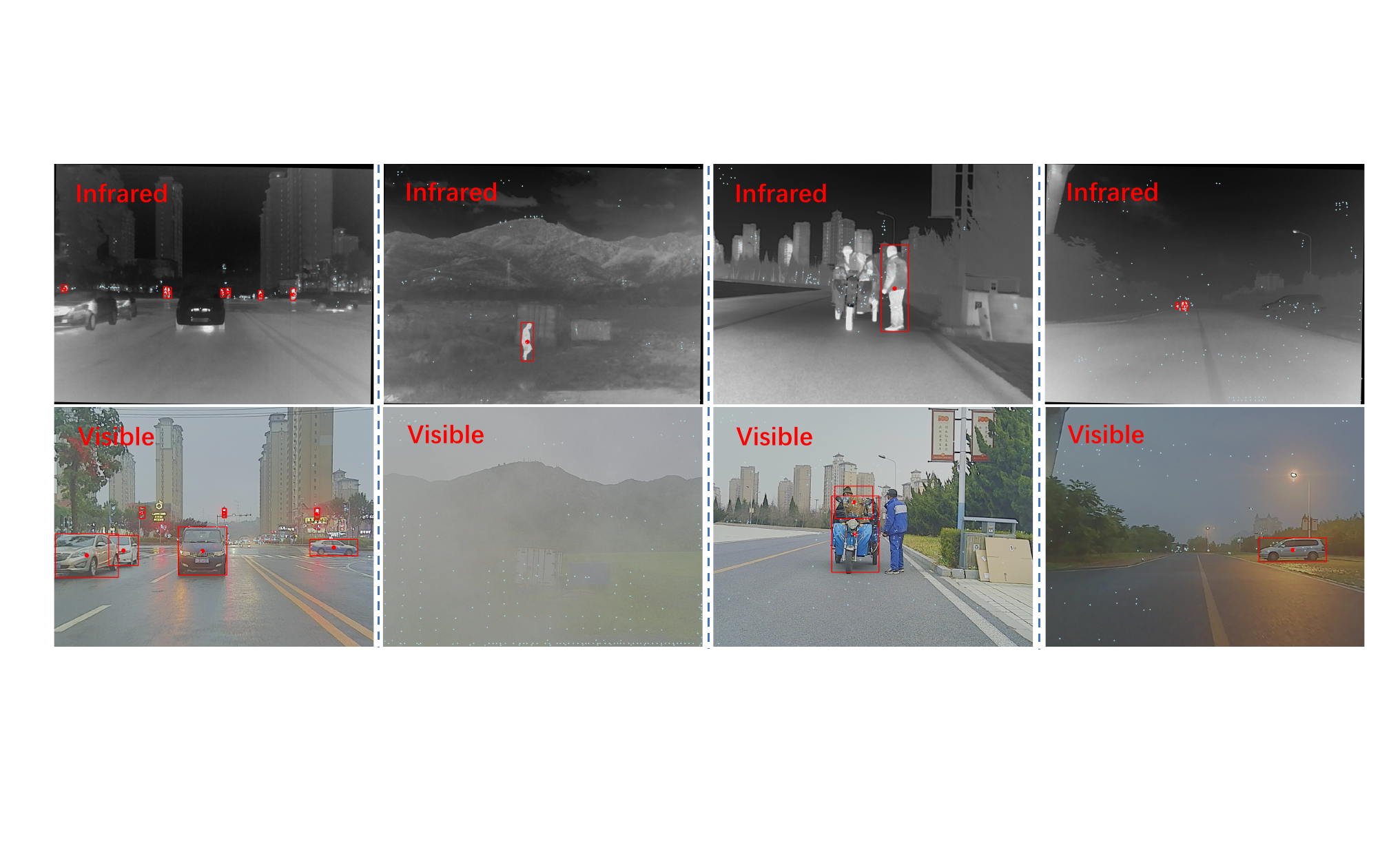}
    \caption{Visualization of Modality Competitive Query Selection results. Red points indicate high-score queries selected in the corresponding modality image, while blue points represent lower-scoring queries. The red boxes indicate the objects represented by high-score queries. }
    \label{fig:visualization_of_modal_competitive_query_selection}
    \vspace{-15pt}
\end{figure}
These Top-$K$ features are sourced either from the infrared or visible features, respectively, and each feature represents an object instance specific to their respective modality.
This approach can be defined as follows:
\vspace{-3pt}
\begin{equation}
    z=\operatorname{Top-\mathit{K}}(\operatorname{Linear}(\operatorname{concat}(I, V))),
\vspace{-5pt}
\end{equation}
where $z$ denotes the set of $K$ selected features, $I$ and $V$ represent the flattened encoded infrared and visible feature sequence, respectively.

As mentioned before, infrared or visible images may contain unuseful interfering information, which could confuse the network.
We competitively select modality-specific features to build a salient feature representation for each object. This approach helps to prevent the introduction of interference from another modality in the early stages and provides useful prior information for refining the query in the subsequent decoder, emphasizing that the query's representation of the object should prioritize the modality from which it originates.
Additionally, we use the optimization strategy of IOU-aware classification loss \cite{zhang2021varifocalnet,lv2023detrs} to further improve the quality of the selected features.

\textbf{Effectiveness analysis.} To observe the performance of our Modality Competitive Query Selection strategy in the network, we visualize the positions and scores of the selected modality-specific initial queries on paired images.
Specifically, we map these selected features to obtain the coordinates of reference points, which are further projected onto either infrared or visible images.
As illustrated in Fig. \ref{fig:visualization_of_modal_competitive_query_selection}, the visualization shows that different object instances are saliently represented by different dominant modality features.
This selection result is consistent with our intuition, demonstrating that this approach dynamically chooses the dominant detection modality for each object under varying conditions.
More detailed quantitative results and analysis are presented in Sec. \ref{ablation_study}.

\textbf{Redundant queries.} We also observe that there are redundant queries in infrared and visible images to point to the same object. However, the network effectively eliminates this redundant information thanks to the DETR's one-to-one matching optimization pattern and the self-attention mechanism applied to all modality-specific queries in the decoder. To further address this potential issue, we introduce Noise Query Learning \cite{zhang2022dino,li2022dn} strategy during training to facilitate learning the optimal modality match for each object. 

\begin{figure}
    \centering
    \includegraphics[trim=0mm 34mm 0mm 34mm,clip,width=0.90\textwidth]{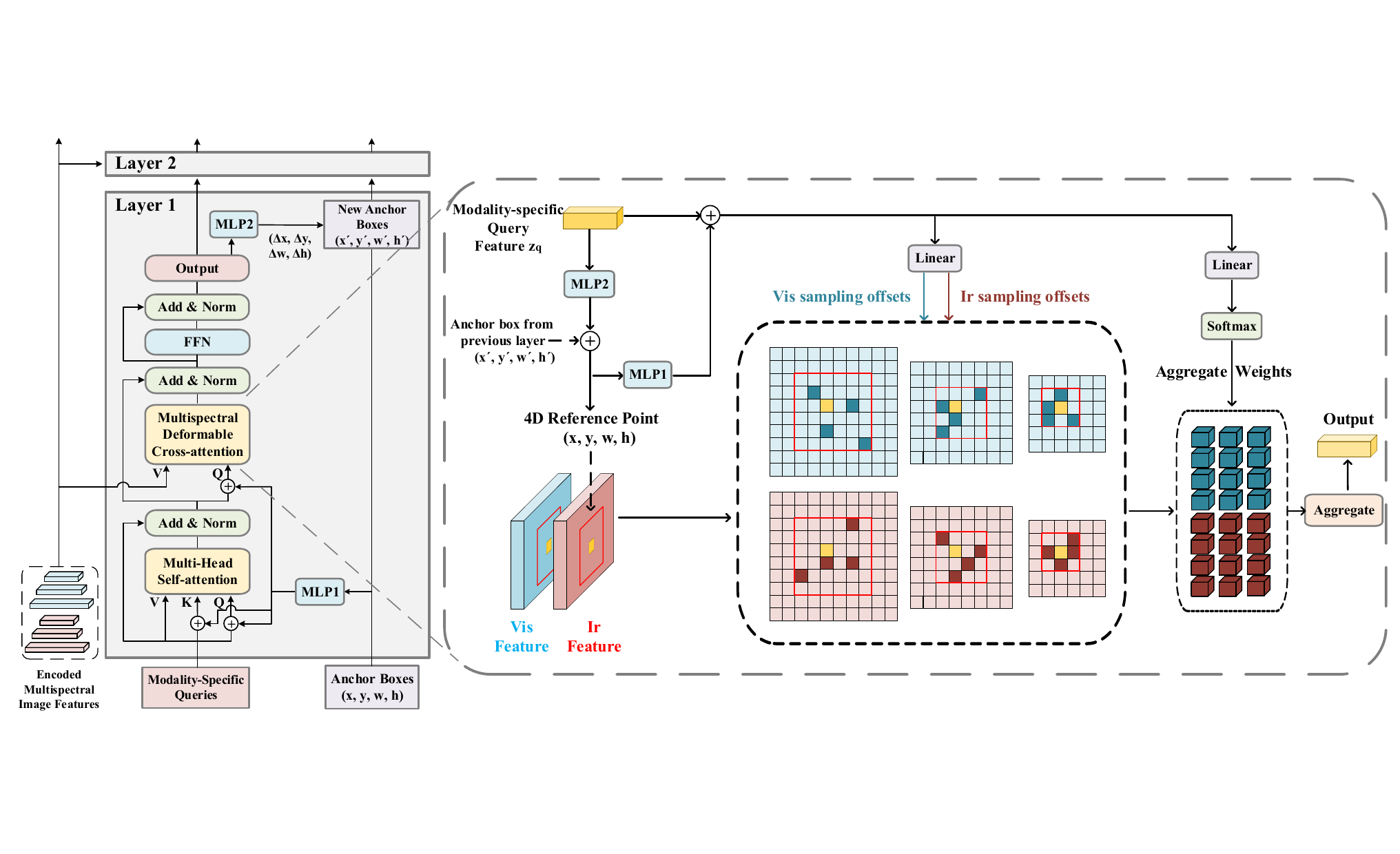}
    \caption{The structure of the Multispectral Transformer Decoder (DeNoising Training Group is omitted in the figure) and Multispectral Deformable cross-attention module. }
    \label{fig:ms-decoder_and_ms-cal}
\vspace{-12pt}
\end{figure}

\vspace{-10pt}
\subsection{Multispectral Transformer Decoder}
\label{decoder}
\quad The details of the Multispectral Transformer Decoder are illustrated in Fig. \ref{fig:ms-decoder_and_ms-cal}.
In each layer, modality-specific object queries first undergo multi-head self-attention to obtain contextual information and reduce redundancy. 
Subsequently, our Multispectral Deformable Cross-attention module refines these modality-specific queries with multi-semantic infrared and visible features.
Additionally, we employ 4D anchor boxes to constrain the sampling range within the Multispectral Deformable Cross-attention module, and iteratively refine queries and anchor boxes through cascaded decoder layers.
Concretely, in the Multispectral Decoder with $D$ layers, we map from the $q$-th modality-specific query $z_{q}^{d}$ in the $d$-th layer to obtain the refined 4D reference point $b_{q}^{d}$. The process can be described as follows:
\vspace{-3pt}
\begin{equation}
    b_{q\{x, y, w, h\}}^{d}=\sigma\left(M L P^{d}\left(z_{q}^{d}\right)+\sigma^{-1}\left(b_{q}^{d-1}\right)\right),
\end{equation}
where $d \in\{2, 3, \ldots, D\}$, $MLP$ consists of two linear projection layers,  $\sigma$ represents 
the sigmoid function, $\sigma^{-1}$ represents the inverse sigmoid function, and $b_{q}^{1}$ is the initialized anchor boxes. The initial anchor setting is consistent with the two-stage method in Deformable DETR \cite{zhu2020deformable}.
This refined 4D reference point serves as the reference position constraint for subsequent sampling on multi-semantic infrared and visible feature maps in the Multispectral Deformable Cross-attention module.

\textbf{Multispectral Deformable Cross-attention module.} In Deformable DETR \cite{zhu2020deformable}, key features are aggregated by sparse sampling on the feature map, and we extend it to a multi-modal form to achieve adaptive infrared and visible feature fusion.
The detailed architecture of the Multispectral Deformable Cross-Attention module is illustrated in Fig. \ref{fig:ms-decoder_and_ms-cal}.
Specifically, we map the 4D reference points to position embeddings through an MLP layer. After combining the modality-specific query feature with position embeddings, two linear layers are employed to predict the sampling offsets and aggregation weights on the multi-semantic feature maps in both modalities, respectively.
\begin{figure}
    \centering
    \includegraphics[trim=2mm 10mm 2mm 10mm,clip,width=\textwidth]{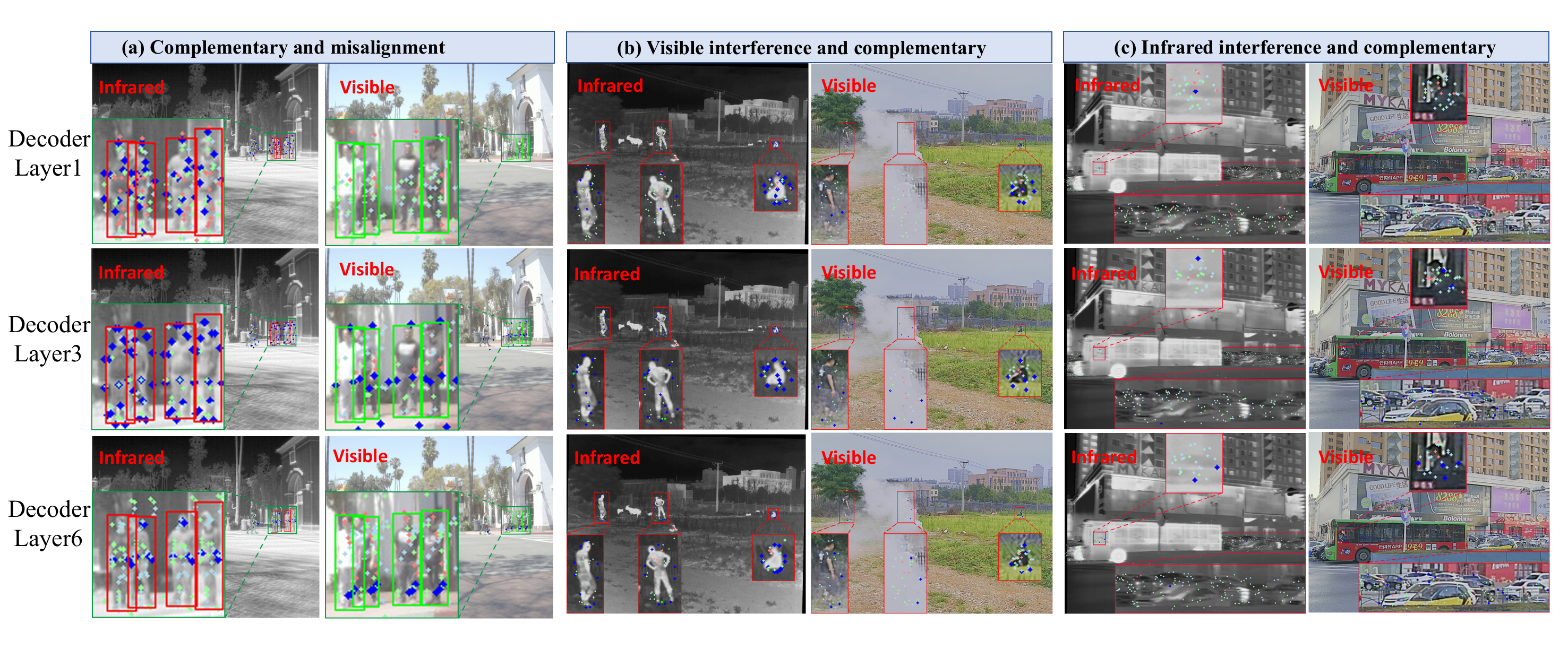}
    \caption{Visualization of feature sampling at different semantic levels in different decoder layers. Different colors of points represent the results of sampling in different semantic layers, where blue, green, and red represent sampling points on low-level, middle-level, and high-level semantic features maps respectively. Brightly colored and large points indicate relatively high attention weights. (a) The green boxes in the visible image represent aligned bounding boxes, which show the sampling points in each modality are concentrated on the right instance locations.  (b) Objects occluded by background and smoke tend to predominantly focus on the infrared modality at subsequent decoder layers.
    (c) Objects in good illumination conditions and those less distinguishable in the infrared modality tend to predominantly focus on the visible modality at subsequent decoder layers. }
    \label{fig:visualization_of_feature_sampling}
    \vspace{-10pt}
\end{figure}
Finally, these sampled multi-semantic infrared and visible features are aggregated by aggregation weights.  
Due to the ability of this method to predict sampling position offsets independently in both the infrared and visible modalities, the network could still focus on the features of misaligned objects in misaligned image pairs.

Given the input multi-semantic infrared and visible feature maps $\left\{x_{1}^{l}, x_{2}^{l}\right\}_{l=1}^{L}$, we use the normalized centerpoint of $b_{q}$ as the 2D reference point $\hat{p}_q$. We define the Multispectral Deformable Cross-attention module $F$ as follows:
\vspace{-8pt}
\begin{equation} 
\begin{split}
    &\mathit { F }\left(z_q, \hat{p}_q,\left\{x_1^l, x_2^l\right\}_{l=1}^L\right)=\sum_h^H W_h \\[-8pt]
    &\left[\sum_m \sum_{l=1}^L \sum_{k=1}^K A_{m h l q k} \cdot W_h^{\prime} x_m^l\left(\Phi_l\left(\hat{p}_q\right)+\Psi_l\left(\Delta \mathrm{p}_{m h l q k}\right)\right)\right],
\end{split}
\end{equation}
where $m \in\{1, 2\}$ denotes the visible and infrared modalities, $h$ indexes the attention head, $l$ indexes the input feature semantic level, and $k$ indexes the sampling point. $A_{m h l q k}$ and $\Delta \mathrm{p}_{m h l q k}$ denote the $k$-th attention weight and sampling point in the $l$-th feature semantic level and the $h$-th attention head within the $m$ modality, respectively. The attention weight $A_{m h l q k}$ is normalized by $\sum_m \sum_{l=1}^L \sum_{k=1}^K A_{m h l q k}=1$. The function $\Phi_l\left(\hat{p}_q\right)$ scales $\hat{p}_q$ to the $l$-th semantic level feature map and function $\Psi_l\left(\Delta \mathrm{p}_{m h l q k}\right)$ constrains the predicted offset within the range of $b_{q}$, so as to focus on the information around the object and reduce the difficulty of optimization.

\textbf{Effectiveness analysis.} To observe the effectiveness of this feature fusion approach, we visualize the positions and weights sampled in both modalities at different semantic levels in different decoder layers, as shown in Fig. \ref{fig:visualization_of_feature_sampling}.
It shows that as the decoder layer deepens, our method tends to adaptively focus on low-level semantic features in infrared modality and additional high-level semantic features in visible modality.
This result is reasonable since the infrared modality carries less information and could provide reliable low-level semantic information such as basic contours and shapes, while the visible modality with more information could additionally provide more abstract high-level semantic information, such as more reliable contextual relationships of object categories.
Additionally,  We observe that these points can adapt to misalignment scenes and adaptively focus on the key information of the object, such as the edge information, which is important to define the boundary of the object.
The differences in sampling positions and weight distribution in different decoder layers also verify the effectiveness of the cascade structure for reliable complementary information mining. More details can be found in Sec. \ref{ablation_study}.

\vspace{-10pt}
\subsection{Loss Function}
\quad The training loss of our model follows the DETR-like detectors, which is defined as:
\vspace{-2pt}
\begin{equation}
\mathcal{L}= \mathcal{L}_{cls}+ \mathcal{L}_{box}+ \mathcal{L}_{d n} ,
\vspace{-2pt}
\end{equation}
where $ \mathcal{L}_{cls}$ is the IoU-aware classification loss following RT-DETR \cite{lv2023detrs}, $\mathcal{L}_{box}$ is composed of L1 loss and generalized IoU loss for bounding box regression and $\mathcal{L}_{d n}$ is the loss for denoising training \cite{zhang2022dino}. In addition, we also calculate the loss of each decoder layer as the auxiliary optimization loss.

\vspace{-15pt}
\section{Experiments}
\label{sec:experiments}
\vspace{-5pt}
\subsection{Dataset and metric}

\quad We conduct experiments on four datasets, which cover objects in different scenarios and scales. We use the standard COCO AP metric as the evaluation metric. The four datasets are M$^3$FD \cite{liu2022target}, FLIR \cite{flir}, LLVIP \cite{jia2021llvip} and VEDAI \cite{razakarivony2016vehicle}.

\textbf{M$^3$FD.} The M$^3$FD dataset contains 4,200 pairs of infrared-visible images with a resolution of 1024 × 768. It covers diverse scenes and six object categories, with slight misalignment of image pairs.  Since the dataset does not provide a public split, we split the dataset into a training set of 3,368 pairs and a validation set containing 831 pairs by different scenes. This implies a low similarity in scenes between the training and validation sets. 

\textbf{FLIR.} We use the aligned version \cite{zhang2020multispectral} with a resolution of 640 × 512, which contains 4,129 pairs for training and 1,013 pairs for testing and has three object categories. This dataset contains both day and night scenes, with obvious misalignment of image pairs.

\textbf{LLVIP.} The dataset is for pedestrian detection in low-light surveillance scenes. It consists of 12,025 image pairs for training and 3,463 pairs for testing with a resolution of 1280 × 1024, with good image pair registration. Most of the scenes are in dark condition, with only one category of pedestrians.

\begin{table}[]
    \centering
    \scriptsize 
\caption{Comparisons on the M$^3$FD Dataset. TarDAL$_{CT}$ represents the infrared-visible fusion image obtained by \cite{liu2022target}. † Includes image fusion and object detection inference time.}
\label{tab:tab1_m3fd_result}
    \begin{tabular}{ccccccc} 
         \toprule
         Model &Backbone& 
      Data Type& $m$AP50& $m$AP75&$m$AP &Inference Speed(s)\\
       \midrule
 Yolov7&CSPDarknet53& IR& 60.5& -&34.8 &0.014\\
 Yolov7&CSPDarknet53& RGB& 69.0& -&42.7 &0.014\\
 Yolov7&CSPDarknet53& TarDAL$_{CT}$ \cite{liu2022target}& 67.0& -&40.0 &0.153†\\
 DINO& ResNet50& IR& 58.8& 36.1&35.0 &0.155\\
 DINO& ResNet50& RGB& 73.3& 48.2&46.3 &0.155\\
 DINO& ResNet50& TarDAL$_{CT}$ \cite{liu2022target}& 68.3& 45.4&43.3 &0.294†\\
 \midrule
 CFT \cite{qingyun2021cross} &CSPDarknet53& IR+RGB& 68.2& 44.6&42.5 &0.050\\
 ICAFusion \cite{shen2024icafusion} &CSPDarknet53& IR+RGB& 67.8& 44.5&41.9 &0.033\\
 Ours &ResNet50& IR+RGB& \textbf{80.2}& \textbf{56.0}&\textbf{52.9} &0.117\\ 
 \bottomrule
 \end{tabular}
\end{table}

\begin{table}[]
    \centering
    \scriptsize 
\caption{Comparisons on the FLIR-aligned Dataset.}
\label{tab:tab2_flir_result}
    \begin{tabular}{cccccc}
         \toprule
         Model& 
      Backbone&Data Type& $m$AP50& $m$AP75&$m$AP\\
     \midrule
 Yolov5&  CSPDarknet53&IR& 80.1& -&42.4\\
 Yolov5 \cite{qingyun2021cross}&  CSPDarknet53&RGB& 67.8& 25.9&31.8\\
 DINO& ResNet50& IR& 80.6& 42.7&44.8\\
 DINO& ResNet50& RGB& 70.9& 25.9&\\
 \midrule
 GAFF \cite{zhang2021guided}&  ResNet18&IR+RGB& 72.9& 32.9&36.6\\
 SMPD \cite{li2023stabilizing}&  VGG16&IR+RGB& 73.6& -&-\\
 CFT \cite{qingyun2021cross}&  CSPDarknet53&IR+RGB& 78.7& 35.5&40.2\\
 ICAFusion \cite{shen2024icafusion}&  CSPDarknet53&IR+RGB& 79.2& 36.9&41.4\\
 MFPT \cite{zhu2023multi}&  ResNet50&IR+RGB& 80.0& -&-\\
 LRAF-Net \cite{fu2023lraf}&  CSPDarknet53&IR+RGB& 80.5& -&42.8\\
 Ours&  ResNet50&IR+RGB& \textbf{86.6}& \textbf{48.1}&\textbf{49.3}\\
 \bottomrule
 \end{tabular}
  \vspace{-5pt}

 \vspace{-10pt}
\end{table}

\textbf{VEDAI.} The dataset is the multispectral aerial imagery dataset for vehicle detection, consisting of  1,200 image pairs with a resolution of 1024 × 1024, and its image pairs are strictly registered. It includes nine object categories and most of the objects are small, which poses a great challenge for DETR-like detectors. We convert its bounding boxes to horizontal boxes, following \cite{qingyun2022cross}.

\vspace{-10pt}
\subsection{Implementation Details}
\quad We employ ResNet50 \cite{he2016deep} as the backbone for both the infrared and visible branches, with a feature map semantic level of L = 3. The Efficient Encoder contains one layer, while the MS-Decoder contains six layers. We set the number of attention heads, sampling points, and selected queries as H = 8, K = 4, and N = 300, respectively. 

We use pre-trained weights on the COCO dataset and employ only basic data augmentations such as random resize, crop, and flip during training.  The learning rate is set to 0.0001 on M$^3$FD, FLIR, and LLVIP, and 0.00025 on VEDAI. 
For fair comparisons with SOTA methods, we set the input image size to 640 × 640 for training and testing on M$^3$FD and FLIR datasets, while 1024 × 1024 for LLVIP and VEDAI datasets. We conducted training for 20 epochs on the FLIR and LLVIP datasets, and for 50 epochs on the M$^3$FD and VEDAI datasets.  Networks were trained using an Nvidia RTX3090 GPU.

\vspace{-15pt}
\subsection{Comparison with SOTA}
\vspace{-5pt}

\quad \textbf{Comparisons on M$^3$FD:}
\begin{table}[]
    \centering
    \scriptsize
\caption{Comparisons on the LLVIP Dataset.}
\label{tab:tab3_llvip_result}
    \begin{tabular}{cccccc}
         \toprule
         Model& 
      Backbone&Data Type& $m$AP50& $m$AP75&$m$AP\\
     \midrule
 Yolov5 \cite{jia2021llvip}&  CSPDarknet53&IR& 96.5& 76.4&67.0\\
 Yolov5 \cite{jia2021llvip}&  CSPDarknet53&RGB& 90.8& 56.4&52.7\\
 DINO& ResNet50& IR& 96.6& 70.6&62.9\\
 DINO& ResNet50& RGB& 91.6& 58.0&53.8\\
 \midrule
 CSAA \cite{cao2023multimodal}&  ResNet50&IR+RGB& 94.3& 66.6&59.2\\
 CFT \cite{qingyun2021cross}&  CSPDarknet53&IR+RGB& 97.5& 72.9&63.6\\
 LRAF-Net \cite{fu2023lraf}&  CSPDarknet53&IR+RGB& 97.9& -&66.3\\
 MS-DETR \cite{xing2023multispectral}&  ResNet50&IR+RGB& 97.9& 76.3&66.1\\
 Ours&  ResNet50&IR+RGB& \textbf{97.9}& \textbf{79.1}&\textbf{69.6}\\
 \bottomrule
 \end{tabular}
  \vspace{-5pt}
\end{table}
\begin{table}[]
    \centering
    \scriptsize
\caption{Comparisons on the VEDAI Dataset.}
\label{tab:tab4_vedai_result}
    \begin{tabular}{ccccc}
         \toprule
         Model& 
      Backbone&Data Type& $m$AP50&$m$AP\\
     \midrule
 Yolo-Fine \cite{pham2020yolo}&  CSPDarknet53&IR& 75.2&-\\
 Yolo-Fine \cite{pham2020yolo}&  CSPDarknet53&RGB& 76.0&-\\
 DINO& ResNet50& IR& 78.4&41.7\\
 DINO& ResNet50& RGB& 81.3&44.9\\
 \midrule
 Yolo Fusion \cite{qingyun2022cross}&  CSPDarknet53&IR+RGB& 78.6&49.1\\
 CFT \cite{qingyun2021cross}&  CSPDarknet53&IR+RGB& 85.3&56.0\\
 ICAFusion \cite{shen2024icafusion}& CSPDarknet53& IR+RGB& 84.8&56.6\\
 LRAF-Net \cite{fu2023lraf}&  CSPDarknet53&IR+RGB& 85.9&\textbf{59.1}\\
 Ours&  ResNet50&IR+RGB& \textbf{91.5}&55.3\\
 \bottomrule
 \end{tabular}
\vspace{-15pt}
\end{table}
The M$^3$FD dataset we split is very challenging with different scenes and all SOTA methods have low values on $m$AP, as shown in Tab. \ref{tab:tab1_m3fd_result}. 
However, our method has much better performance and obviously outperforms these SOTA methods. Specifically,  our method outperforms CFT \cite{qingyun2021cross} by 12\% on $m$AP50, 11.4\% on $m$AP75, and 10.4\% on $m$AP. It also outperforms ICAFusion \cite{shen2024icafusion} by 12.4\% on $m$AP50, 11.5\% on $m$AP75, and 11\% on $m$AP. This significant performance improvement shows that our method can better adapt to complex and changeable scenes in infrared-visible object detection. 

\textbf{Comparisons on FLIR:} As shown in Tab. \ref{tab:tab2_flir_result}, our method outperforms single-modality methods and surpasses the best feature fusion methods by 6.1\% on $m$AP50 and 6.5\% on overall $m$AP, demonstrating that our method can effectively mine complementary information under different illumination conditions.

\textbf{Comparisons on LLVIP:} As shown in Tab. \ref{tab:tab3_llvip_result}, the dataset has nearly saturated performance on $m$AP50, but our method still outperforms the best of $m$AP75 (MS-DETR \cite{xing2023multispectral}) by 2.8\%, and the best of $m$AP (LRAF-Net \cite{fu2023lraf}) by 3.3\%, demonstrating our method could mine more comprehensive and fine-grained complementary information. 

\textbf{Comparisons on VEDAI:} As shown in Tab. \ref{tab:tab4_vedai_result}, our method still can work well for small objects, and outperforms most of the SOTA methods. Especially,  on $m$AP50, our method outperforms the best (LRAF-Net \cite{fu2023lraf}) by 5.6\%. 
On $m$AP, our method does not outperform LRAF-Net \cite{fu2023lraf} which is a CNN-based detector. The reason could be that in comparison to long-range information extraction, small object detection more relies on local feature extraction, and CNNs could be more advantageous than Transformer in this regard. On the other hand, the smaller bounding boxes of these small objects are sensitive to displacements, and even a few pixel displacements will lead to significant changes in IOU and subsequent $m$AP reduction.
\renewcommand{\floatpagefraction}{.9}
\begin{figure}
    \vspace{-8pt}
    \centering
    \includegraphics[trim=2mm 8mm 8mm 8mm,clip,width=\textwidth]{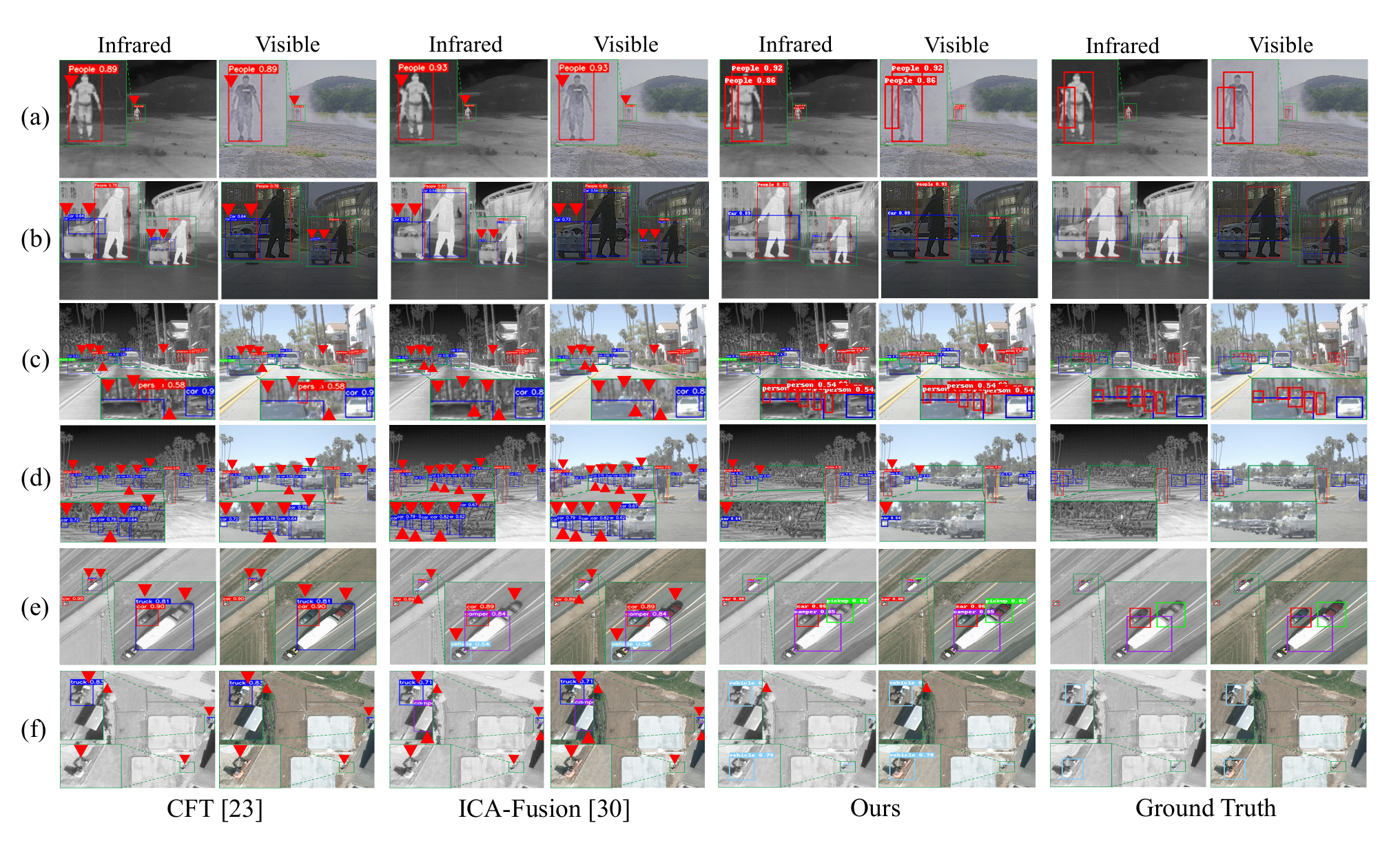}
    \vspace{-17pt}
    \caption{Representative results on the M$^3$FD(a, b), FLIR(c, d) and VEDAI(e, f). The inverted red triangles indicate detection results that do not match the ground truth. The confidence threshold is set to 0.5 when visualizing these results.
    }
    \label{fig:detection_example}
    \vspace{-15pt}
\end{figure}
Therefore, high precision in bounding box prediction is crucial for small object detection. Relatively, CNN methods can achieve higher bounding box accuracy, resulting in a higher $m$AP.
Nevertheless, the improvement in $m$AP50 demonstrates the potential of our method in small object scenes.

\textbf{Detection Visualization.} For quality analysis, we provide some representative detection results on M$^3$FD(a, b), FLIR(c, d) and VEDAI(e, f) datasets. As shown in Fig. \ref{fig:detection_example}, our method can accurately locate objects and achieve higher detection confidence in different scences. These results demonstrate that our DAMSDet can adaptively focus on dominant modalities and effectively mine fine-grained multi-level semantic complementary information.


\vspace{-10pt}
\subsection{Ablation Study on M$^3$FD}
\label{ablation_study}

\quad To verify the effectiveness of key modules and strategies of our method, we perform ablation experiments on the high-quality, multi-scenes M$^3$FD dataset. 

\textbf{Effect of Modality Competitive Query Selection (MCQS).} We apply standard query selection on the added encoded features to show the effectiveness of  Modality Competitive Query Selection. As shown in the 5\textsuperscript{th} and 6\textsuperscript{th} rows of the Tab. \ref{tab:tab5_ablation_result},  our Modality Competitive Query Selection strategy bring 1.1\% $m$AP50 and 0.7\% $m$AP improvement due to avoiding the early introduction of interference by dynamically selecting modality-specific queries.

\textbf{Effect of Multispectral deformable cross-attention module (MDCA).}  We apply standard deformable cross-attention to the fused features obtained by adding the outputs of two encoders in the decoder. 
As shown in the 5\textsuperscript{th} and 7\textsuperscript{th} rows of the Tab. \ref{tab:tab5_ablation_result}, our method brings improvements of 1.5\% $m$AP50 and 0.9\% $m$AP. 
\begin{table}[]
    \centering
    \scriptsize
    
\caption{Ablation studies on the M$^3$FD dataset. MCQS, MDCA, and CQS represent Modality Competitive Query Selection, Multispectral Deformable Cross-attention, and content query selection respectively.}
\label{tab:tab5_ablation_result}
    \begin{tabular}{c|cc|cc|c|ccc}
         \toprule
          \multirow{2}{*}{} & \multicolumn{2}{c|}{Modality} & \multicolumn{3}{c|}{Module \& Strategy} & \multicolumn{3}{c}{Metrics} \\
        \cmidrule{2-9}
   &Vis& Ir& MCQS& MDCA&CQS& $m$AP50& $m$AP75&$m$AP\\
  \midrule
   1&$\checkmark$& & & & $\checkmark$& 75.7& 51.9&49.3\\
  2&$\checkmark$& & & & & 75.3& 52.0&49.4\\
  3&& $\checkmark$& & & $\checkmark$& 62.2& 40.1&38.5\\
  4&& $\checkmark$& & & & 62.5& 39.9&39.0\\
 \midrule
   5&$\checkmark$& $\checkmark$& & &$ \checkmark$& 77.8& 56.0&51.6\\
   6&$\checkmark$& $\checkmark$& $\checkmark$& & $\checkmark$& 78.9& 55.5&52.3\\
   7&$\checkmark$& $\checkmark$& & $\checkmark$& $\checkmark$& 79.4& 55.8&52.5\\
   8&$\checkmark$& $\checkmark$& $\checkmark$& $\checkmark$& & 79.3& 55.8&52.6\\
   9&$\checkmark$& $\checkmark$& $\checkmark$& $\checkmark$& $\checkmark$& \textbf{80.2}& \textbf{56.0}&\textbf{52.9}\\
  \bottomrule
 \end{tabular}
 
\vspace{-15pt}
\end{table}
This improvement can be attributed to our method's ability of effectively adapting to modality misalignment scenes while performing adaptive feature aggregation for each object in a more fine-grained way.
Certainly, when using both Modality Competitive Query Selection and the Multispectral Deformable Cross-Attention module, we achieved the best results on all AP metrics, as shown in the 9\textsuperscript{th} row of the Tab. \ref{tab:tab5_ablation_result}.

\textbf{Effect of Content Query Selection (CQS).} We analyze the performance of different content query strategies in infrared-visible object detection. 
For comparison, we take the single branch of our network as the single-modality detection method. 
In the 8\textsuperscript{th} and 9\textsuperscript{th} rows of the Tab. \ref{tab:tab5_ablation_result}, without CQS means setting the content query as a learnable query, while the position query is still obtained by Modality Competitive Query Selection (similar to mixed query selection in DINO \cite{zhang2022dino}). 
We can see that using CQS leads to better results, as it provides stronger prior information for subsequent multispectral cross-attention in the Multispectral Decoder. However, as shown in the first 4 rows of Tab. \ref{tab:tab5_ablation_result}, the selection or learnability of content query has a limited impact on single-modality object detection, since the feature representation of single modality is relatively consistent.

\vspace{-10pt}
\subsection{Limitation}
\quad 
Our method is effective in addressing common misalignment in the majority of cases. However, it may not handle extreme misalignment situations well, as the object features of the modality will be lost when objects exceed the range of the 4D reference point.  

\vspace{-10pt}
\section{Conclusion}
\label{sec:conclusion}
\vspace{-5pt}
\quad In this paper, we propose DAMSDet to simultaneously address complementary information fusion and modality misalignment problems in infrared-visible object detection. Through Modality Competitive Query Selection, DAMSDet can dynamically choose salient modality feature representations for specific objects based on complementary characteristics. In the Multispectral Deformable Cross-attention module, we link feature fusion and modality misalignment to mine reliable complementary information at multi-semantic levels. Experiments on four datasets with different scenarios demonstrate that the proposed method achieves significant improvements compared to other state-of-the-art methods.


%
%
\bibliographystyle{splncs04}
\bibliography{main}

\begin{thebibliography}{10}
\providecommand{\url}[1]{\texttt{#1}}
\providecommand{\urlprefix}{URL }
\providecommand{\doi}[1]{https://doi.org/#1}

\bibitem{cao2019box}
Cao, Y., Guan, D., Wu, Y., Yang, J., Cao, Y., Yang, M.Y.: Box-level segmentation supervised deep neural networks for accurate and real-time multispectral pedestrian detection. ISPRS journal of photogrammetry and remote sensing  \textbf{150},  70--79 (2019)

\bibitem{cao2023multimodal}
Cao, Y., Bin, J., Hamari, J., Blasch, E., Liu, Z.: Multimodal object detection by channel switching and spatial attention. In: Proceedings of the IEEE/CVF Conference on Computer Vision and Pattern Recognition. pp. 403--411 (2023)

\bibitem{carion2020end}
Carion, N., Massa, F., Synnaeve, G., Usunier, N., Kirillov, A., Zagoruyko, S.: End-to-end object detection with transformers. In: European conference on computer vision. pp. 213--229. Springer (2020)

\bibitem{flir}
FLIR: Flir thermal dataset for algorithm training. \url{https://www.flir.in/oem/adas/adas-dataset-form} (2018)

\bibitem{fu2023lraf}
Fu, H., Wang, S., Duan, P., Xiao, C., Dian, R., Li, S., Li, Z.: Lraf-net: Long-range attention fusion network for visible--infrared object detection. IEEE Transactions on Neural Networks and Learning Systems  (2023)

\bibitem{he2016deep}
He, K., Zhang, X., Ren, S., Sun, J.: Deep residual learning for image recognition. In: Proceedings of the IEEE conference on computer vision and pattern recognition. pp. 770--778 (2016)

\bibitem{jia2021llvip}
Jia, X., Zhu, C., Li, M., Tang, W., Zhou, W.: Llvip: A visible-infrared paired dataset for low-light vision. In: Proceedings of the IEEE/CVF international conference on computer vision. pp. 3496--3504 (2021)

\bibitem{kim2021mlpd}
Kim, J., Kim, H., Kim, T., Kim, N., Choi, Y.: Mlpd: Multi-label pedestrian detector in multispectral domain. IEEE Robotics and Automation Letters  \textbf{6}(4),  7846--7853 (2021)

\bibitem{kim2021uncertainty}
Kim, J.U., Park, S., Ro, Y.M.: Uncertainty-guided cross-modal learning for robust multispectral pedestrian detection. IEEE Transactions on Circuits and Systems for Video Technology  \textbf{32}(3),  1510--1523 (2021)

\bibitem{konig2017fully}
Konig, D., Adam, M., Jarvers, C., Layher, G., Neumann, H., Teutsch, M.: Fully convolutional region proposal networks for multispectral person detection. In: Proceedings of the IEEE conference on computer vision and pattern recognition workshops. pp. 49--56 (2017)

\bibitem{li2018multispectral}
Li, C., Song, D., Tong, R., Tang, M.: Multispectral pedestrian detection via simultaneous detection and segmentation. arXiv preprint arXiv:1808.04818  (2018)

\bibitem{li2019illumination}
Li, C., Song, D., Tong, R., Tang, M.: Illumination-aware faster r-cnn for robust multispectral pedestrian detection. Pattern Recognition  \textbf{85},  161--171 (2019)

\bibitem{li2022dn}
Li, F., Zhang, H., Liu, S., Guo, J., Ni, L.M., Zhang, L.: Dn-detr: Accelerate detr training by introducing query denoising. In: Proceedings of the IEEE/CVF Conference on Computer Vision and Pattern Recognition. pp. 13619--13627 (2022)

\bibitem{li2022confidence}
Li, Q., Zhang, C., Hu, Q., Fu, H., Zhu, P.: Confidence-aware fusion using dempster-shafer theory for multispectral pedestrian detection. IEEE Transactions on Multimedia  (2022)

\bibitem{li2023stabilizing}
Li, Q., Zhang, C., Hu, Q., Zhu, P., Fu, H., Chen, L.: Stabilizing multispectral pedestrian detection with evidential hybrid fusion. IEEE Transactions on Circuits and Systems for Video Technology  (2023)

\bibitem{liu2016multispectral}
Liu, J., Zhang, S., Wang, S., Metaxas, D.N.: Multispectral deep neural networks for pedestrian detection. arXiv preprint arXiv:1611.02644  (2016)

\bibitem{liu2022target}
Liu, J., Fan, X., Huang, Z., Wu, G., Liu, R., Zhong, W., Luo, Z.: Target-aware dual adversarial learning and a multi-scenario multi-modality benchmark to fuse infrared and visible for object detection. In: Proceedings of the IEEE/CVF Conference on Computer Vision and Pattern Recognition. pp. 5802--5811 (2022)

\bibitem{liu2022dab}
Liu, S., Li, F., Zhang, H., Yang, X., Qi, X., Su, H., Zhu, J., Zhang, L.: Dab-detr: Dynamic anchor boxes are better queries for detr. arXiv preprint arXiv:2201.12329  (2022)

\bibitem{lv2023detrs}
Lv, W., Xu, S., Zhao, Y., Wang, G., Wei, J., Cui, C., Du, Y., Dang, Q., Liu, Y.: Detrs beat yolos on real-time object detection. arXiv preprint arXiv:2304.08069  (2023)

\bibitem{meng2021conditional}
Meng, D., Chen, X., Fan, Z., Zeng, G., Li, H., Yuan, Y., Sun, L., Wang, J.: Conditional detr for fast training convergence. In: Proceedings of the IEEE/CVF International Conference on Computer Vision. pp. 3651--3660 (2021)

\bibitem{pham2020yolo}
Pham, M.T., Courtrai, L., Friguet, C., Lef{\`e}vre, S., Baussard, A.: Yolo-fine: One-stage detector of small objects under various backgrounds in remote sensing images. Remote Sensing  \textbf{12}(15), ~2501 (2020)

\bibitem{qingyun2021cross}
Qingyun, F., Dapeng, H., Zhaokui, W.: Cross-modality fusion transformer for multispectral object detection. arXiv preprint arXiv:2111.00273  (2021)

\bibitem{qingyun2022cross}
Qingyun, F., Zhaokui, W.: Cross-modality attentive feature fusion for object detection in multispectral remote sensing imagery. Pattern Recognition  \textbf{130},  108786 (2022)

\bibitem{razakarivony2016vehicle}
Razakarivony, S., Jurie, F.: Vehicle detection in aerial imagery: A small target detection benchmark. Journal of Visual Communication and Image Representation  \textbf{34},  187--203 (2016)

\bibitem{redmon2016you}
Redmon, J., Divvala, S., Girshick, R., Farhadi, A.: You only look once: Unified, real-time object detection. In: Proceedings of the IEEE conference on computer vision and pattern recognition. pp. 779--788 (2016)

\bibitem{redmon2017yolo9000}
Redmon, J., Farhadi, A.: Yolo9000: better, faster, stronger. In: Proceedings of the IEEE conference on computer vision and pattern recognition. pp. 7263--7271 (2017)

\bibitem{redmon2018yolov3}
Redmon, J., Farhadi, A.: Yolov3: An incremental improvement. arXiv preprint arXiv:1804.02767  (2018)

\bibitem{ren2015faster}
Ren, S., He, K., Girshick, R., Sun, J.: Faster r-cnn: Towards real-time object detection with region proposal networks. Advances in neural information processing systems  \textbf{28} (2015)

\bibitem{roszyk2022adopting}
Roszyk, K., Nowicki, M.R., Skrzypczy{\'n}ski, P.: Adopting the yolov4 architecture for low-latency multispectral pedestrian detection in autonomous driving. Sensors  \textbf{22}(3), ~1082 (2022)

\bibitem{shen2024icafusion}
Shen, J., Chen, Y., Liu, Y., Zuo, X., Fan, H., Yang, W.: Icafusion: Iterative cross-attention guided feature fusion for multispectral object detection. Pattern Recognition  \textbf{145},  109913 (2024)

\bibitem{wang2023yolov7}
Wang, C.Y., Bochkovskiy, A., Liao, H.Y.M.: Yolov7: Trainable bag-of-freebies sets new state-of-the-art for real-time object detectors. In: Proceedings of the IEEE/CVF Conference on Computer Vision and Pattern Recognition. pp. 7464--7475 (2023)

\bibitem{xing2023multispectral}
Xing, Y., Wang, S., Liang, G., Li, Q., Zhang, X., Zhang, S., Zhang, Y.: Multispectral pedestrian detection via reference box constrained cross attention and modality balanced optimization. arXiv preprint arXiv:2302.00290  (2023)

\bibitem{yang2022baanet}
Yang, X., Qian, Y., Zhu, H., Wang, C., Yang, M.: Baanet: Learning bi-directional adaptive attention gates for multispectral pedestrian detection. In: 2022 International Conference on Robotics and Automation (ICRA). pp. 2920--2926. IEEE (2022)

\bibitem{yao2021efficient}
Yao, Z., Ai, J., Li, B., Zhang, C.: Efficient detr: improving end-to-end object detector with dense prior. arXiv preprint arXiv:2104.01318  (2021)

\bibitem{zhang2022dino}
Zhang, H., Li, F., Liu, S., Zhang, L., Su, H., Zhu, J., Ni, L.M., Shum, H.Y.: Dino: Detr with improved denoising anchor boxes for end-to-end object detection. arXiv preprint arXiv:2203.03605  (2022)

\bibitem{zhang2021varifocalnet}
Zhang, H., Wang, Y., Dayoub, F., Sunderhauf, N.: Varifocalnet: An iou-aware dense object detector. In: Proceedings of the IEEE/CVF conference on computer vision and pattern recognition. pp. 8514--8523 (2021)

\bibitem{zhang2020multispectral}
Zhang, H., Fromont, E., Lefevre, S., Avignon, B.: Multispectral fusion for object detection with cyclic fuse-and-refine blocks. In: 2020 IEEE International conference on image processing (ICIP). pp. 276--280. IEEE (2020)

\bibitem{zhang2021guided}
Zhang, H., Fromont, E., Lef{\`e}vre, S., Avignon, B.: Guided attentive feature fusion for multispectral pedestrian detection. In: Proceedings of the IEEE/CVF winter conference on applications of computer vision. pp. 72--80 (2021)

\bibitem{zhang2021weakly}
Zhang, L., Liu, Z., Zhu, X., Song, Z., Yang, X., Lei, Z., Qiao, H.: Weakly aligned feature fusion for multimodal object detection. IEEE Transactions on Neural Networks and Learning Systems  (2021)

\bibitem{zhang2019weakly}
Zhang, L., Zhu, X., Chen, X., Yang, X., Lei, Z., Liu, Z.: Weakly aligned cross-modal learning for multispectral pedestrian detection. In: Proceedings of the IEEE/CVF international conference on computer vision. pp. 5127--5137 (2019)

\bibitem{zhou2020improving}
Zhou, K., Chen, L., Cao, X.: Improving multispectral pedestrian detection by addressing modality imbalance problems. In: Computer Vision--ECCV 2020: 16th European Conference, Glasgow, UK, August 23--28, 2020, Proceedings, Part XVIII 16. pp. 787--803. Springer (2020)

\bibitem{zhu2020deformable}
Zhu, X., Su, W., Lu, L., Li, B., Wang, X., Dai, J.: Deformable detr: Deformable transformers for end-to-end object detection. arXiv preprint arXiv:2010.04159  (2020)

\bibitem{zhu2023multi}
Zhu, Y., Sun, X., Wang, M., Huang, H.: Multi-modal feature pyramid transformer for rgb-infrared object detection. IEEE Transactions on Intelligent Transportation Systems  (2023)

\end{thebibliography}
\end{document}